\documentclass[10pt,twocolumn,letterpaper]{article}

\usepackage{cvpr}
\usepackage{times}
\usepackage{epsfig}
\usepackage{graphicx}
\usepackage{amsmath}
\usepackage{amssymb}

\usepackage{algorithm}
\usepackage{algorithmic}
\usepackage{array}
\renewcommand{\algorithmicrequire}{\textbf{Input:}}
\renewcommand{\algorithmicensure}{\textbf{Output:}}

\usepackage{url}

\cvprfinalcopy 

\def\cvprPaperID{****} 

\begin{document}

\title{Adaptive Substring Extraction and Modified Local NBNN Scoring for Binary Feature-based Local Mobile Visual Search without False Positives}

\author{Yusuke Uchida$^{\dagger, \ddagger}$ \\
$\dagger$The University of Tokyo\\
Tokyo, Japan\\
\and
Shigeyuki Sakazawa$^{\ddagger}$ \\
$\ddagger$KDDI R\&D Laboratories, Inc. \\
Saitama, Japan
\and
Shin'ichi Satoh$^{\dagger, \dagger\dagger}$ \\
$\dagger\dagger$National Institute of Informatics \\
Tokyo, Japan
}

\maketitle

\begin{abstract}
In this paper, we propose a stand-alone mobile visual search system based on binary features and the bag-of-visual words framework.
The contribution of this study is three-fold: (1) We propose an adaptive substring extraction method that adaptively extracts informative bits from the original binary vector and stores them in the inverted index. These substrings are used to refine visual word-based matching. (2) A modified local NBNN scoring method is proposed in the context of image retrieval, which considers the density of binary features in scoring each feature matching. (3) In order to suppress false positives, we introduce a convexity check step that imposes a convexity constraint on the configuration of a transformed reference image.
The proposed system improves retrieval accuracy by 11\% compared with a conventional method without increasing the database size.
Furthermore, our system with the convexity check does not lead to false positive results.
\end{abstract}


\section{Introduction}
\label{sec:intro}
With the advances in both stable interest region detectors~\cite{mik_ijcv05} and robust and distinctive descriptors~\cite{mik_pami05}, local feature-based image or object retrieval has become a popular research topic.
In particular, binary local features such as Oriented FAST and Rotated BRIEF (ORB)~\cite{rub_iccv11} have attracted much attention due to their efficiency.
Binary features are one or two orders of magnitude faster than Scale Invariant Feature Transform (SIFT)~\cite{low04} or Speeded Up Robust Features (SURF)~\cite{bay_cviu08} features in detection and description, while providing comparable performance~\cite{rub_iccv11, hei_eccv12}.
These binary features are especially suitable for mobile visual search or augmented reality on mobile devices\cite{yan_ismar12}.

With the increasingly widespread use of mobile devices such as Android phones or iPhones, mobile visual search (MVS) has become one of the major applications of image retrieval and recognition technology.
While some research focuses on server-client systems in the context of MVS, the purpose of our research is to achieve fast and accurate recognition with lower memory requirements on mobile devices~\cite{pan_iccv13, dav_mm14}; in this paper, we call the latter type of MVS "local MVS".
Local MVS does not require any server and it works without a network, achieving faster recognition.
Thus, it is suitable for recognizing medium sized databases:
i.e., recognizing catalogs, paintings in a museum, or cards in a collectible card game.
As these kinds of objects can be considered as explicitly or approximately planar~\cite{phi07}, we focus on recognizing planar objects in this study.
We also assume a real-time local MVS system or application, where images captured by a mobile device's camera are continuously used as an input to the local MVS system.
If the camera captures an object registered in the database, our system is expected to show information or content related to the object immediately.
This use case further requires local MVS systems to suppress undesirable false positives because objects in the database do not always appear in captured images.

The difficulty for the local MVS lies in the indexing of local features because it is necessary to fit the database into the size of the memory in mobile devices or an application while maintaining retrieval accuracy.
In other words, managing the trade-off between the memory size of the database and the accuracy of image retrieval is very important.
We assume that the proposed local MVS system is implemented as a mobile application, and the application is provided via digital distribution services such as Play Store or App Store.
In such a situation, the application size is very important because users tend not to download large-size applications.
In this study, we aim at maximizing the accuracy under the constraint of a reasonably small application size (e.g. no larger than 10 MB).

When indexing binary features, Locality Sensitive Hashing (LSH)~\cite{gio_vldb99} is often used~\cite{rub_iccv11, yan_ismar12}.
However, this does not satisfy our constraint because it requires a large amount of memory;
original feature vectors and many hash tables must be stored in LSH~\cite{jeg10, muj_crv12}.

A Bag-of-Visual Words (BoVW) framework~\cite{siv03} is the most widely-used approach for local feature-based image or object retrieval that achieves fast retrieval with lower memory requirements.
As there is room for improvement in the accuracy of the standard BoVW framework, many methods have been proposed to improve this framework for continuous features such as SIFT~\cite{phi07, phi_cvpr08, jeg_ijcv10, mik_eccv10}.
However, there are not many studies on indexing binary features for the purpose of image retrieval.
In \cite{gal_iros11}, the BoVW framework has been adopted to index recent binary features, referred to as Bag-of-Binary Words (BoBW).
In \cite{zho_mm12}, a variant of the Hamming embedding method~\cite{jeg_ijcv10} is proposed for binarized features in order to improve the trade-off between memory requirements and accuracy.
In this method, the quantization process is performed by treating the first $a$ bits of a binary feature as an integer ranging from 0 to $2^a-1$.
Such VWs are not robust because two binary features that are different only in one bit out of the first $a$ bits are quantized into different VWs.

In this study, in order to achieve a real-time local MVS system, we propose a variant of the BoBW framework, which adaptively extracts and stores VW-dependent information of each binary feature to refine VW-based matching.
As the scoring method for matched feature pairs has not been considered in depth and only the standard tf-idf scoring is used in \cite{zho_mm12}, we also propose a modified version of the local Naive Bayes Nearest Neighbor (local NBNN) scoring method for image retrieval, which was originally proposed for image classification~\cite{mcc_cvpr12}.
It provides a theoretical basis for scoring feature matching in voting and the proposed modification improves performance by using adaptive density estimation without any additional overhead.
Finally, we introduce a geometric verification method in order to suppress false positives.
This paper is the extended version of the paper~\cite{uch_gcce14} that appeared in Global Conference on Consumer Electronics (GCCE) 2014.
In particular, we introduce a new geometric verification method and show that our system can achieve a zero false positive rate.

In this study, we did not consider the Fisher vector approach~\cite{per_eccv10, jeg_pami12, uch_acpr13} or the Vector of Locally Aggregated Descriptors (VLAD) approach~\cite{jeg_cvpr10, ara_cvpr13, dav_mm14, spy_tmm14}, which achieves reasonable retrieval accuracy with very compact image representation.
However geometric verification becomes unavailable in these approaches because matching pairs of features are not obtained in the search process.
Geometric verification is essential to ensure a low false positive rate as also shown in our experiment later.
Although it is possible to perform feature-level matching after an image-level search, this makes these approaches inefficient and reduces their advantages.

In summary, the contributions of this research are three-fold, as follows:
\begin{enumerate}
\item We propose an adaptive substring extraction method that adaptively extracts informative bits from the original binary vector and stores them in an inverted index. These substrings are used to refine visual word-based matching.
\item A modified local NBNN scoring method is proposed in the context of image retrieval, which considers the density of binary features in scoring each feature matching.
\item In order to suppress false positives, we introduce a convexity check step that imposes a convexity constraint on the configuration of a transformed reference image.
\end{enumerate}

The rest of this paper is organized as follows.
In Section 2, the binary features we are going to use in our system are briefly introduced.
In Section 3, we describe the BoVW framework and its extensions.
In Section 4, our proposed framework is introduced.
In Section 5, the effectiveness of the proposed framework is confirmed.
Our conclusions are presented in Section 6.

\section{Local binary features}
To date, many binary features are proposed such as ORB~\cite{rub_iccv11}, Fast Retina Keypoint (FREAK)~\cite{ala_cvpr12}, Binary Robust Invariant Scalable Keypoints (BRISK)~\cite{leu_iccv11}, KAZE features~\cite{alc_eccv12}, Accelerated-KAZE (A-KAZE)~\cite{alc_bmvc13}, Local Difference Binary (LDB)~\cite{yan_pami14}, and Learned Arrangements of Three patCH codes (LATCH)~\cite{lev_wacv16}.
In this section, binary features are briefly introduced focusing on the ORB feature, which is one of the most frequently used binary features.
The algorithm of a local (binary) feature is divided into two major parts: detection and description.
In detection, local patches are detected in an image, and then binary feature vectors are extracted from these patches in the description.

\subsection{Detection}
Most of the local binary features employ fast feature detectors.
The ORB feature utilizes the Features from the Accelerated Segment Test (FAST)~\cite{ros_iccv05} detector, which detects pixels that are brighter or darker than neighboring pixels based on the accelerated segment test.
The test is optimized to reject candidate pixels very quickly, achieving extremely fast feature detection.
In order to ensure approximate scale invariance, feature points are detected from an image pyramid.

\subsection{Description}
Local binary features extract binary strings from patches of interest regions instead of extracting gradient-based high-dimensional feature vectors like SIFT.
The BRIEF descriptor~\cite{cal_eccv10}, a pioneering work in the area of binary descriptors, is a bit string description of an image patch constructed from a set of binary intensity tests.
Consider the $t$-th smoothed image patch $p_t$, a binary test $\tau$ for $d$-th bit is defined by:
\begin{equation}
	x_{td} = \tau(p_t; a_d, b_d) =
	\begin{cases}
		\, 1 & \mathrm{if} \; p_t(a_d) \ge p_t(b_d) \\
		\, 0 & \mathrm{else}
	\end{cases},
\end{equation}
where $a_d$ and $b_d$ denote relative positions in the patch $p_t$, and $p_t(\cdot)$ denotes the intensity at the point.
Using $D$ independent tests, we obtain $D$-bit binary string $x_t = (x_{t1}, \cdots, x_{td}, \cdots, x_{tD})$ for the patch $p_t$.
The ORB feature employs a learning method for de-correlating BRIEF features under rotational invariance.

\section{Bag-of-visual words and its extensions}
Indexing features is an essential component of efficient retrieval or recognition.
The most widely adopted framework is the BoVW framework~\cite{siv03}.
In this section, the BoVW framework and its extensions are described.

\subsection{Bag-of-visual words framework}
In the BoVW framework, extracted local features of an image are quantized into visual words (VWs), resulting in a histogram representation of VWs.
Here, VWs are representative feature vectors, which are created beforehand by applying the $k$-means algorithm to the training feature vectors.
In many cases, image similarity is measured by the $\ell_1$ or $\ell_2$ distance between the normalized histograms of VWs.
As the histograms are generally sparse, an inverted index and a voting function enable an efficient similarity search~\cite{siv03,jeg_ijcv10}.

\subsection{Extended inverted index}
\label{sec:extended}
Though the BoVW framework achieves efficient retrieval with lower memory requirements, degradation of accuracy is caused by quantization.
In the BoVW framework, two features are matched if and only if they are assigned to the same VW~\cite{jeg_ijcv10}.
Therefore, quite different features are sometimes matched in the BoVW framework.
One approach to suppress these unreliable feature matches is to increase the number of VWs (e.g. 1M).
Although this approach is used in local MVS~\cite{pan_iccv13, dav_mm14} as well as large-scale image retrieval~\cite{nis06, phi07, mik_ijcv13}, it is not suitable for our purpose because it increases the size of the application.
For example, the VWs used in \cite{pan_iccv13} or \cite{dav_mm14} requires 17 MB or 72 MB storage respectively.

The other approach to refine feature matching is the post-filtering approach~\cite{jeg_ijcv10, jeg10} where relatively small size of VWs can be used.
In this approach, after VW-based matching, the distances between a query feature and the reference features that are assigned to the same VW are calculated and matches with large distances are filtered out.
In order to perform post-filtering, the inverted index is extended to store additional information for reference features.
As exact distance calculation is undesirable in terms of the computational cost and memory requirement to store raw feature vectors, short code-based methods are used for this purpose~\cite{jeg_ijcv10, jeg10};
original feature vectors are encoded into short codes and the distances between feature vectors are approximated by the distances between the short codes.
In \cite{jeg_ijcv10}, a pioneer work of this approach, the Hamming embedding (HE) method is proposed.
In HE, local feature vectors of reference images are encoded into binary codes by random projection and thresholding after quantization, and the resulting binary codes are stored in the inverted index.
In the search step, each query feature vector is also binarized after quantization, and the Hamming distances between the binary code of the query feature and the binary codes in the corresponding list of the inverted index are calculated.
Then, scores are voted for the reference images associated with the binary codes whose Hamming distances to the binary code of the query feature are no more than a threshold.
In \cite{jeg_ijcv10}, weak geometric consistency is also proposed that filters matching pairs that are not consistent in terms of angle and scale.
In \cite{jeg10}, instead of the binarization of vectors, a product quantization (PQ) is used to encode reference feature vectors into short codes, and the resulting short codes are also stored in inverted index.

\subsection{Problems of conventional systems}
Although many methods have been proposed to improve the BoVW framework for continuous features, there are not many studies on indexing binary features for the purpose of image retrieval.
In \cite{gal_iros11}, the BoVW framework has been adopted to indexing recent binary features, namely BoBW.
In \cite{zho_mm12}, a variant of the Hamming embedding method is proposed for binarized features, where the first $a$ bits are used to form $2^a$ VWs by treating a bit string as an integer, and the next $b$-bit substring is stored in an inverted index.
However, this approach results in non-optimal performance because directly using the binary string as a VW is not robust against disturbance;
if even one of the first $a$ bits is different between the reference and query binary features, they are quantized into different VWs and never match.

\begin{figure*}[tb]
	\centering
	\begin{minipage}[c]{0.3\linewidth}
		\includegraphics[width=\linewidth]{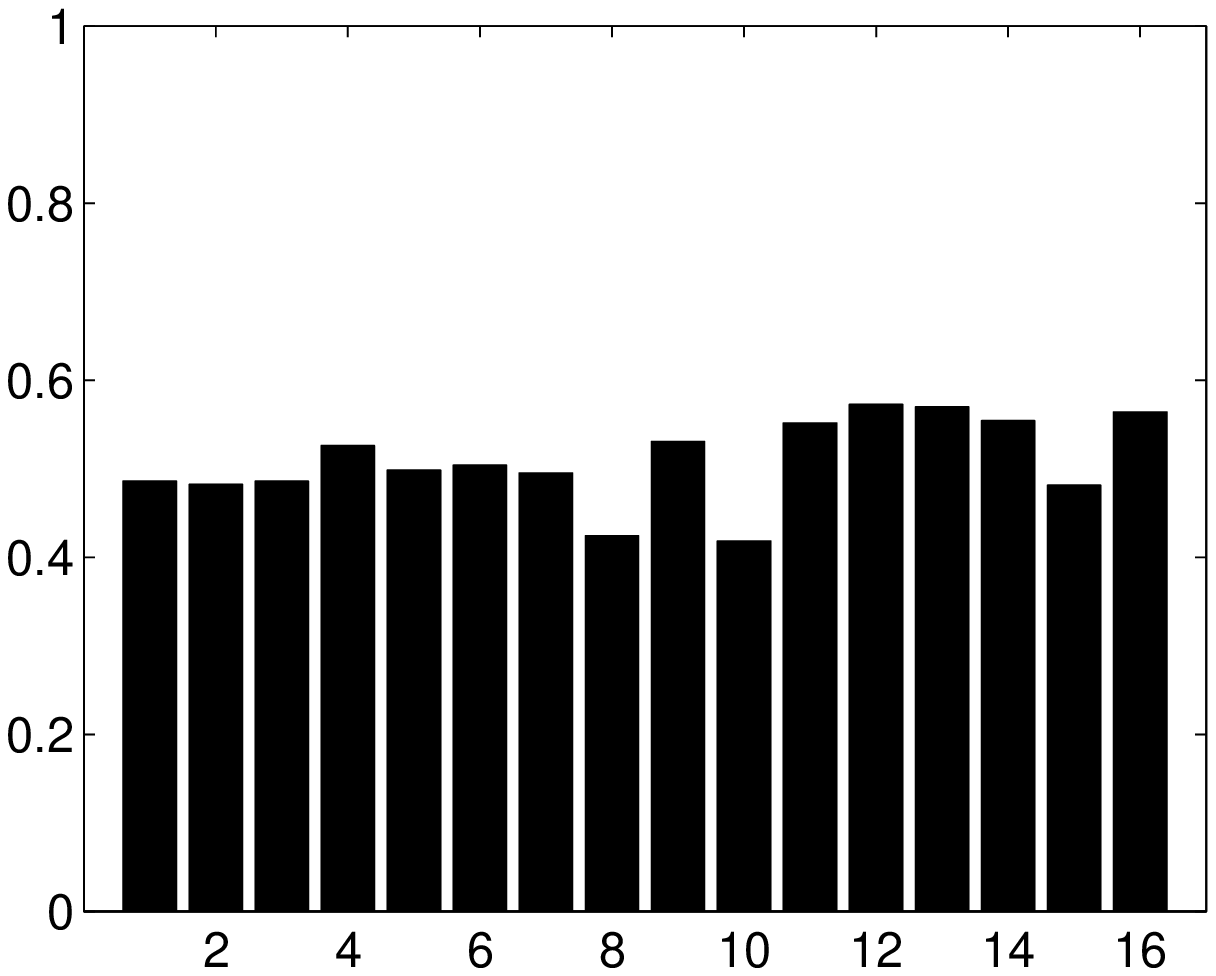} \\
		\centering (a) \\
	\end{minipage}
	\begin{minipage}[c]{0.3\linewidth}
		\includegraphics[width=\linewidth]{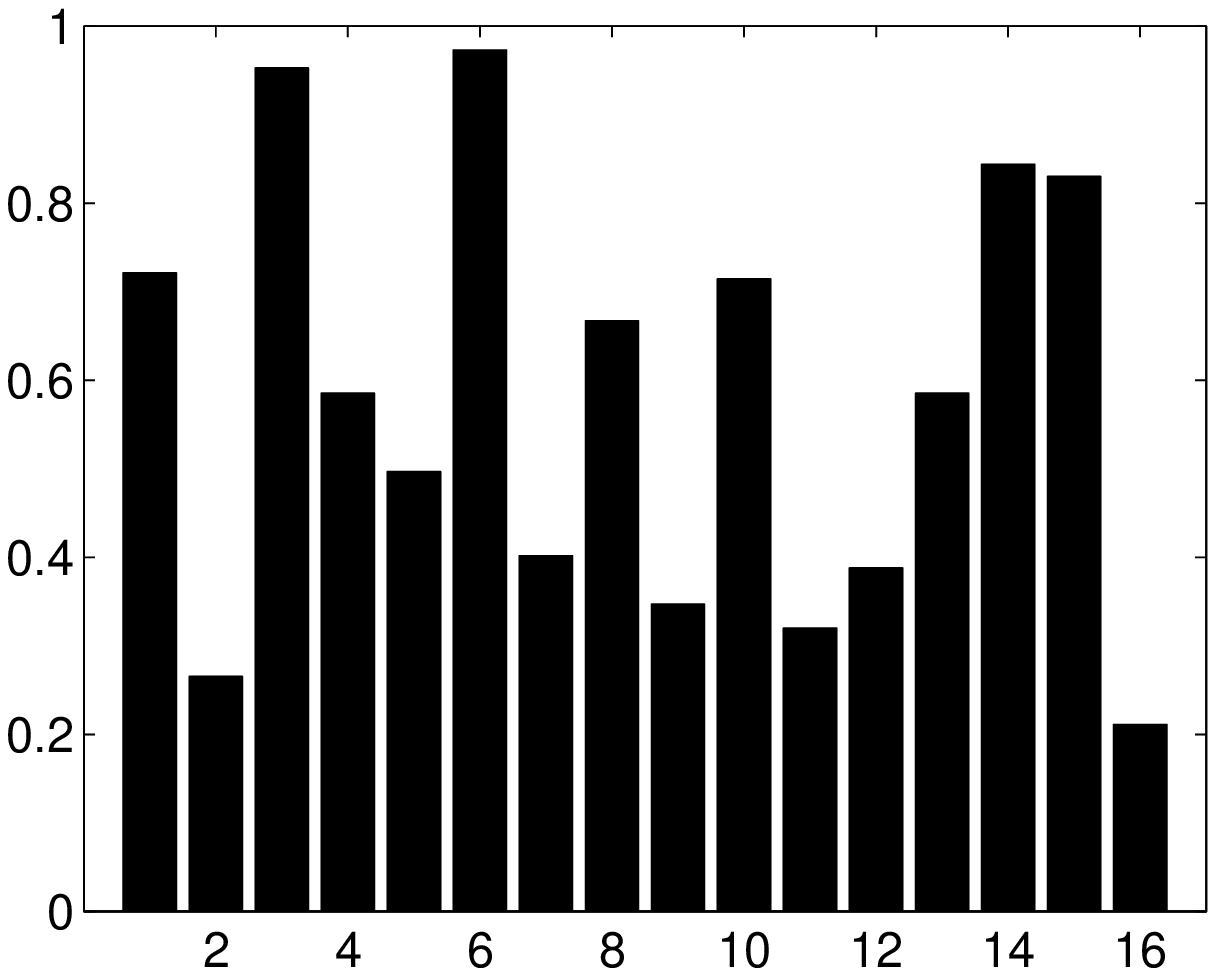} \\
		\centering (b) \\
	\end{minipage}
	\begin{minipage}[c]{0.3\linewidth}
		\includegraphics[width=\linewidth]{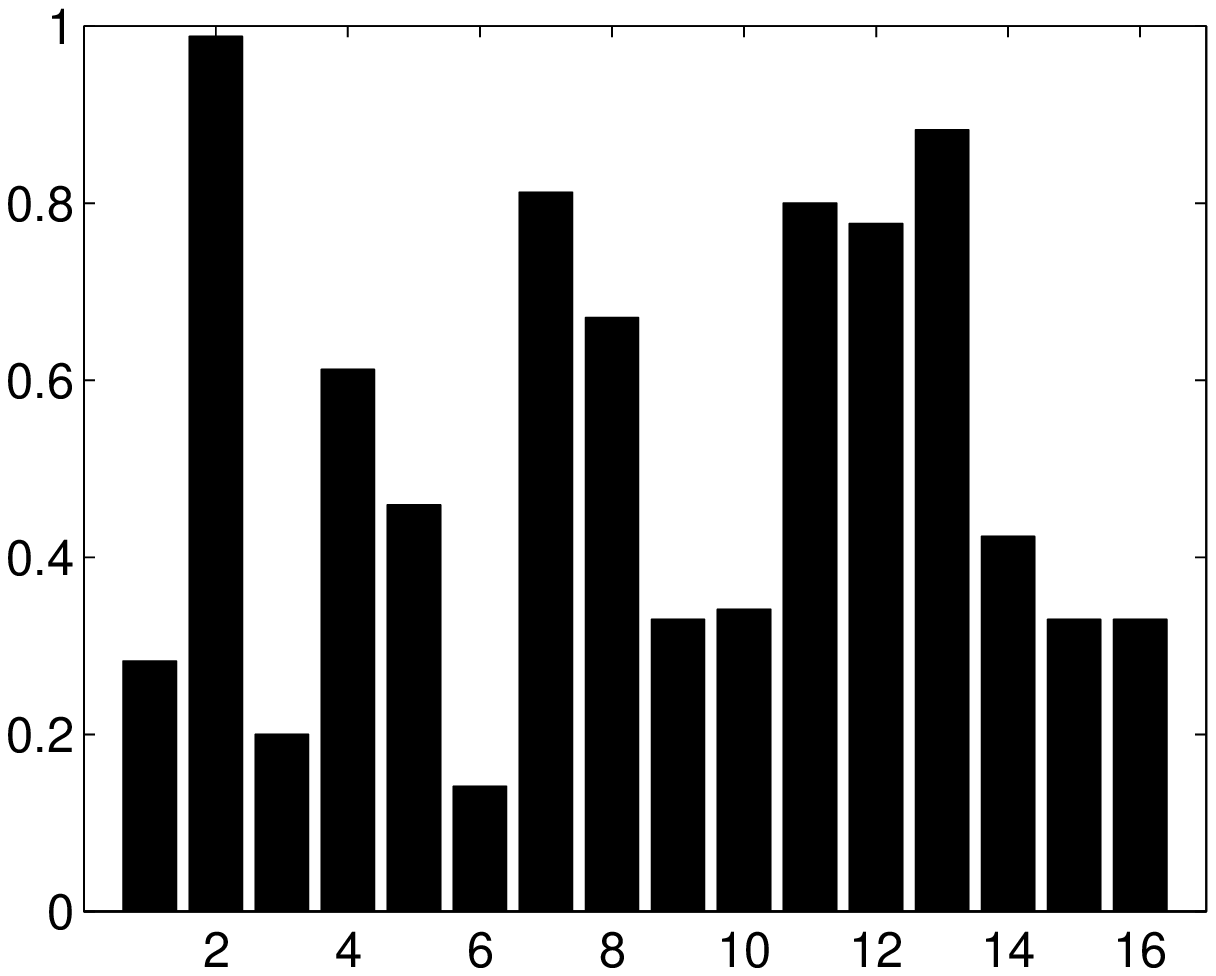} \\
		\centering (c) \\
	\end{minipage} \\
	\begin{minipage}[c]{0.3\linewidth}
		\includegraphics[width=\linewidth]{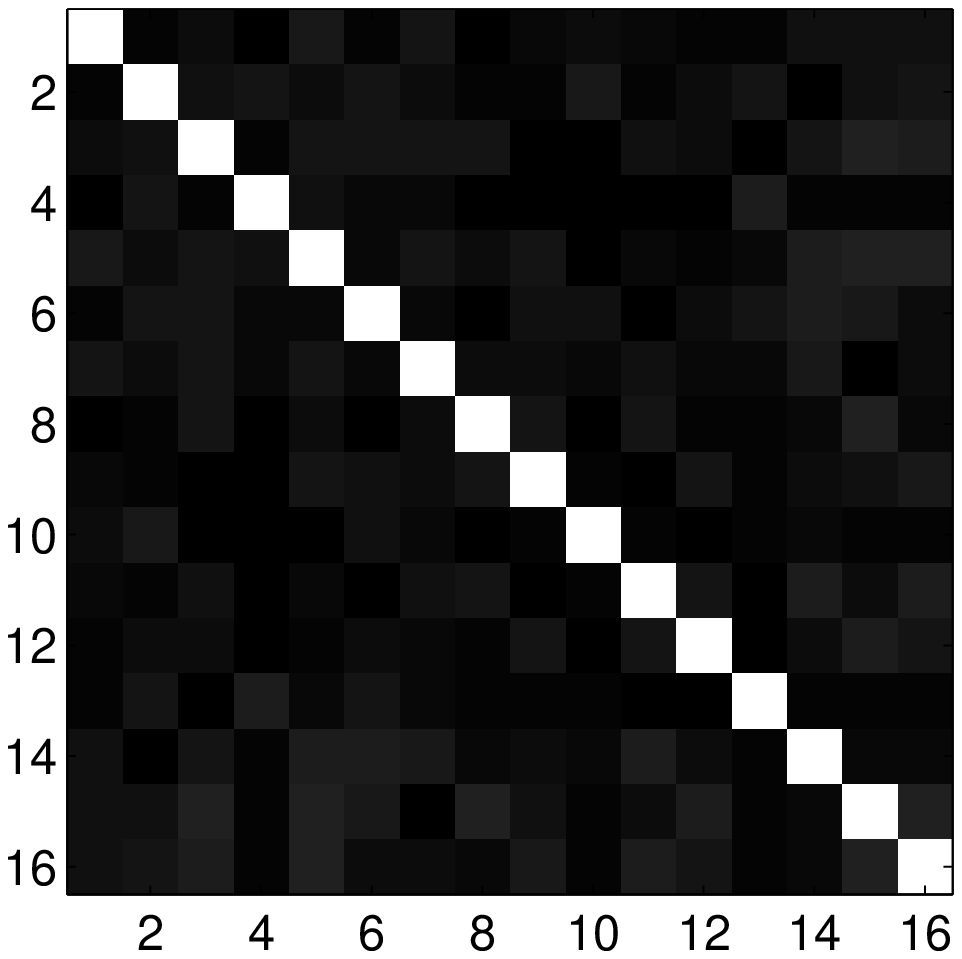} \\
		\centering (d) \\
	\end{minipage}
	\begin{minipage}[c]{0.3\linewidth}
		\includegraphics[width=\linewidth]{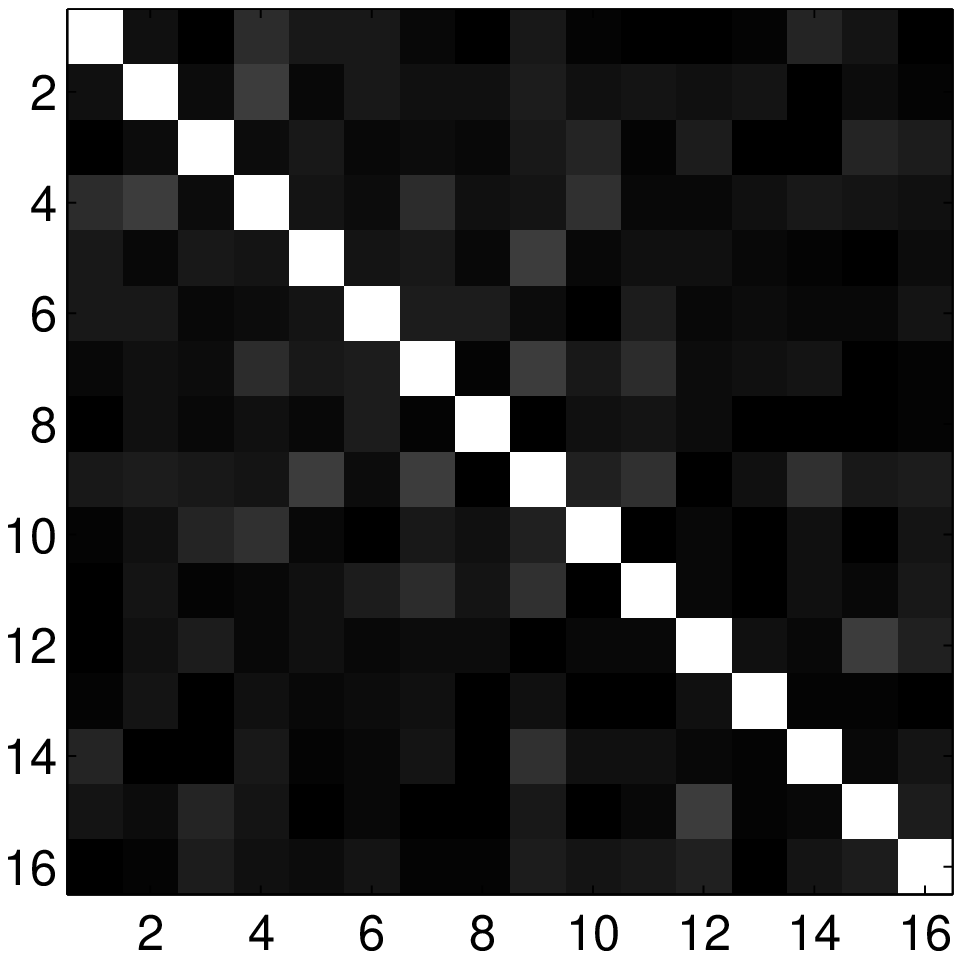} \\
		\centering (e) \\
	\end{minipage}
	\begin{minipage}[c]{0.3\linewidth}
		\includegraphics[width=\linewidth]{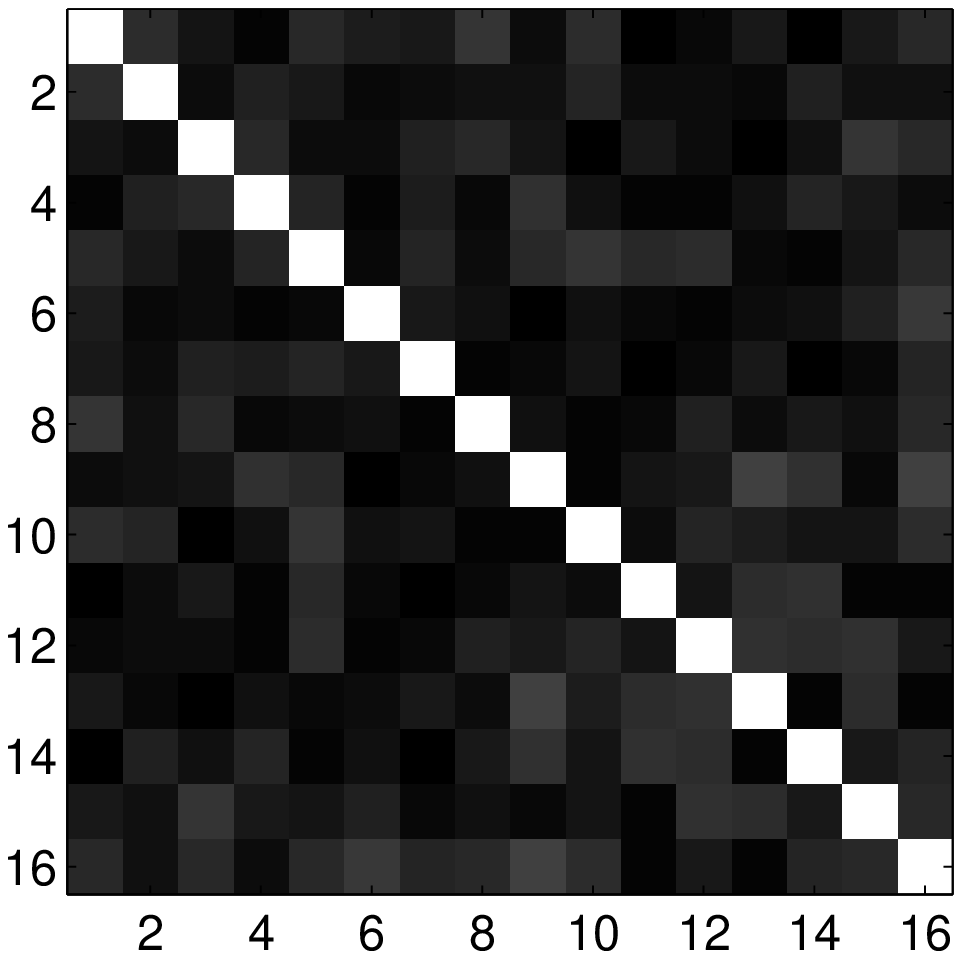} \\
		\centering (f) \\
	\end{minipage} \\
	\caption{(a) and (d) represent mean values and absolute correlation coefficients of first 16 bits of the ORB feature vectors \textit{before} clustering.
		(b) and (e) represent mean values and absolute correlation coefficients of the ORB feature vectors assigned to a certain VW among 1024 VWs, and (c) and (f) correspond to the other VW.}
	\label{fig:corr}
\end{figure*}

The other problem is that non-informative bits are sometimes selected as a substring.
This problem is caused by VW-based clustering.
Figure~\ref{fig:corr} illustrates this problem by considering the statistics of binary features before and after VW-based clustering.
Figure~\ref{fig:corr} (a) and (d) represent mean values and absolute correlation coefficients of the first 16 bits of the ORB feature vectors \textit{before} clustering.
We can see that mean values are close to 0.5 and correlation is weak as designed.
Figure~\ref{fig:corr} (b) and (e) represent mean values and absolute correlation coefficients of the ORB feature vectors assigned to a certain VW among 1024 VWs, and Figure~\ref{fig:corr} (c) and (f) correspond to the other VW.
Figure~\ref{fig:corr} gives us three observations about the statistics \textit{after} VW-based clustering;
(1) the mean values of binary feature vectors are significantly different from 0.5 in some dimensions,
(2) some bits of the binary feature are correlated with each other, and
(3) the characteristics of these correlations and mean values are different from one VW to another.
Thus, using the fixed positions of bits as a substring among VWs is not appropriate.
In the next section, we propose selecting informative bits as a substring and store the substring in an inverted index for post-filtering to solve this problem.

\begin{figure}[tb]
	\centering
	\includegraphics[width=\linewidth]{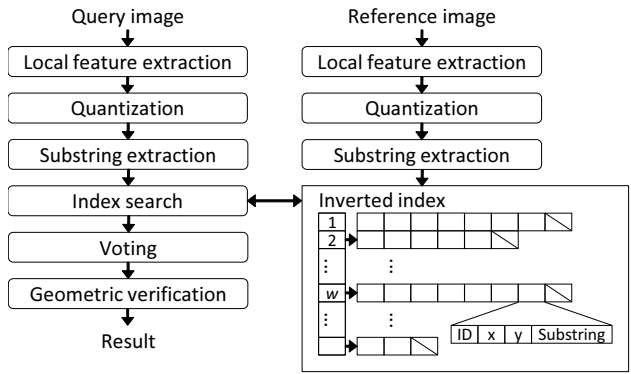}
	\caption{The framework of the proposed method.}
	\label{fig:framework}
\end{figure}

\section{Proposed local MVS system}
In this section, we propose a local MVS system using a post-filtering approach, which is suitable for binary features.
Figure~\ref{fig:framework} shows the framework of the proposed method.
In the indexing step (offline), binary reference features are extracted from reference images and quantized into VWs.
The substrings of reference features are generated and stored in an inverted index to facilitate an efficient search.
In the search step (online), each query feature of a query image votes on reference images with scores according to the distances between the query feature substring and reference feature substrings.
Finally, geometric verification is performed to suppress false positives.

As mentioned in Section~\ref{sec:intro}, our novelty lies in adaptive substring extraction (Section~\ref{sec:extract}), modified local NBNN scoring (Section~\ref{sec:scoring}), and geometric verification with a convexity check (Section~\ref{sec:gv}).

\subsection{Constructing VWs and quantization}
As the proposed system is constructed on the BoVW framework, VWs should be constructed before indexing.
For VW construction, the $k$-means algorithm is performed on the training binary features.
Then, the resulting centroid vectors are binarized by thresholding at 0.5 as done in \cite{gal_iros11}, obtaining binary VWs.
Quantization is done by calculating the Hamming distances between the reference feature and the VWs, and assigning the reference feature to the nearest VW.

\subsection{Adaptive substring extraction}
\label{sec:extract}
While the method proposed in \cite{zho_mm12} extracts the substring using the fixed positions of bits, we extract the substring adaptively changing the bit positions according to the assigned VW.
To this end, we utilize \textit{substring dictionary} $\{ \mathcal{D}_w \}_{w = 1}^{N}$ where $\mathcal{D}_w$ defines the positions of $T$ useful bits for the $w$-th VW.
Letting $w$ denotes the identifier of the nearest VW of the input binary vector $x$, we extract the substring $(x_{\mathcal{D}_{w1}}, \cdots, x_{\mathcal{D}_{wT}})$ from the original binary vector $x = (x_1, \cdots, t_D)$ using $\mathcal{D}_w$.
For example, in the case that $T = 4$ and $\mathcal{D}_w = (4, 25, 70, 87)$, the resulting 4-bit substring becomes $(x_4, x_{25}, x_{70}, x_{87})$.

The substring dictionary $\{ \mathcal{D}_w \}_{w = 1}^{N}$ is constructed with the algorithm used in ORB~\cite{rub_iccv11}, where informative (mean value is close to 0.5) and non-correlated bits are selected.
To do this, training vectors are first clustered into $N$ sets $\{ \mathcal{X}_w \}_{w = 1}^{N}$ using VWs, where $N$ denotes the number of VWs.
Then, for each $w$-th VW, the substring dictionary $\mathcal{D}_w$ is constructed using the training vectors $\mathcal{X}_w$. 
Algorithm~\ref{alg:construct} describes the algorithm for the construction of the substring dictionary $\mathcal{D}_w$ for the $w$-th VW.
In this algorithm, bit positions are first sorted according to their entropy.
Then, each bit position is added to the dictionary $\mathcal{D}_w$ if the bit is not correlated with all of the bit positions already in $\mathcal{D}_w$.
If the algorithm is finished before $T$ bits are selected, the threshold $th$ is decreased and Algorithm~\ref{alg:construct} is performed again.
The resulting dictionary $\{ \mathcal{D}_w \}_{w = 1}^{N}$ is used in the following indexing and search step.
Storing the dictionary requires $N{\times}T$ bytes because one byte can represent one bit position of a 256-bit string.

\begin{algorithm}[tb]
\caption{Substring dictionary construction}
\label{alg:construct}
\begin{algorithmic}[1]
\REQUIRE Training vectors $\mathcal{X}_w$ assinged to $w$-th VW, threshold for correlation $th$ 
\ENSURE Substring dictionary $\mathcal{D}_w$ with size $T$
\STATE $C \leftarrow$ correlation matrix of $\mathcal{X}_w$
\STATE $A \leftarrow$ bit identifiers sorted in ascending order of the absolute difference between the mean value of the bit from 0.5
\STATE $\mathcal{D}_w \leftarrow \{ A_1 \}$
\FOR{$i = 2$ \TO $D$}
	\STATE $j \leftarrow A_i$
	\IF{$\max_{k \in \mathcal{D}_w} |C_{jk}| < th$}
		\STATE $\mathcal{D}_w \leftarrow \mathcal{D}_w \cup j$
		\IF{$|\mathcal{D}_w| = T$}
			\STATE \textbf{break}
		\ENDIF
	\ENDIF
\ENDFOR
\end{algorithmic}
\end{algorithm}

\subsection{Indexing reference images}
In the indexing step, binary features are extracted from reference images and these features are stored in the inverted index as follows.
First, each reference binary feature is quantized into VW.
Letting $w$ denote the identifier of the nearest VW, the substring is extracted using $\mathcal{D}_w$.
Then, the following information is stored in the $w$-th list of the inverted index as shown in Figure~\ref{fig:framework}:
image identifier (2 bytes), the position $(x, y)$ (2+2 bytes), and the substring ($T/8$ bytes).
In total, $6+T/8$ bytes per feature are required.

\subsection{Searching inverted index}
In the search step, binary features are extracted from a query image.
Each query feature votes on reference images with scores using the following procedure.
First, the binary query feature is quantized into VW $w$.
The substring of the binary query feature is generated in the same manner as in the indexing step using $\mathcal{D}_w$.
Then, the distances between the query substring and reference substrings in the corresponding $w$-th list of the inverted index are calculated.
Finally, scores are assigned to the $K$-nearest neighbor reference features.

\subsection{Modified local NBNN scoring}
\label{sec:scoring}
It is known that weighting scores according to their distances improves performance.
The most common way of doing this weighting is to use the Gaussian function $\exp(-d^2/\sigma^2)$~\cite{phi_cvpr08, jeg_cvpr09, jeg_ijcv10}, where $d$ is the Euclidean or Hamming distance between the query feature and reference feature and $\sigma$ is an adjustable parameter.
However, this approach has little theoretical basis and is not optimal.
In this paper, we propose a modified version of the local NBNN (LN) method~\cite{mcc_cvpr12}, which has a theoretical background in the derivation of its score.
Although LN was originally proposed for image classification~\cite{mcc_cvpr12}, we show that this method also works well in image retrieval.
In LN, for each query $q$, its $K$ nearest neighbor features are searched for, where $K$ is an adjustable parameter that specifies the number of samples used in kernel density estimation.
Then, a score of $d^2_K - d^2_k$ is assigned to the corresponding image of the $k$-th nearest neighbor feature, where $d^2_x$ represents the distance between $q$ and its $x$-th nearest neighbor feature.
We refer to this original LN scoring as \textsf{LNo}.
In this study, we modify \textsf{LNo} to $(d_K / d_k)^2 - 1$.
This modification has the effect of adaptively changing the kernel radius in kernel density estimation similar to local scaling~\cite{zel_nips04}, resulting in more appropriate scoring.
We refer to this modified LN scoring method as \textsf{LNm}.

While the parameter $K$ is set to 10 in \cite{mcc_cvpr12}, we empirically use $K = 2$ in this study.
This is because, in the classification task \cite{mcc_cvpr12}, many training images (e.g. 15 or 30) are available for each class, while only a single reference image can be used to represent a reference object in many retrieval tasks.
In classification task, using relatively large $K$ contributes to accuracy because it can utilize multiple training images in density estimation.
In a small-scale retrieval task, using the first and second nearest neighbors is enough, similar to the ratio test done in feature matching~\cite{low04}.

\subsection{Geometric verification with convexity check}
\label{sec:gv}
Geometric Verification (GV) or spatial re-ranking is an important step to improve the results obtained by the voting function~\cite{phi07, chu_iccv07}.
In this step, transformations between the query image and the top-$R$ reference images in the list of voting results are estimated, eliminating matching pairs that are not consistent with the estimated transformation.
In the estimation, the RANdom SAmple Consensus (RANSAC) algorithm or its variants~\cite{chum_accv04, chu_cvpr05} are used.
Then, the score is updated counting only inlier pairs.
As a transformation model, an affine or homography matrix is usually used.
In our case, we estimate the homography matrix using the PROgressive SAmpling and Consensus (PROSAC) algorithm~\cite{chu_cvpr05} for efficiency.
As matching pairs between the query image and the top-$R$ reference images have already been obtained in the voting step, these matching pairs are used as an input to RANSAC.
In many studies, GV is used only for re-ranking~\cite{phi07, chu_iccv07} to improve retrieval accuracy.
In our case, the purpose of GV is to suppress false positives by thresholding the number of inliers.
For this purpose, standard GV is not sufficient;
there is an adequate number of inliers for non-relevant image pairs as will be shown in Section~\ref{sec:gvresult}.

In this study, after standard GV, we check the estimated model under the assumption that reference images represent planar objects.
In particular, we check the \textit{convexity} of the reference image projected to the query image using the Homography matrix estimated in geometric verification.
Let $a$, $b$, $c$, and $d$ denote the four corners of a reference image in clockwise order.
These points are transformed by the estimated homography $H$:
$a' = H a$, $b'= H b$, $c' = H c$, and $d' = H d$.
Here, $a'$, $b'$, $c'$, and $d'$ represent the corners of the reference image captured in the query image.
The angle of each corner of the transformed reference image does not become larger than 180 degrees if the estimated Homography is correct because the transformed reference image should be convex.
We can verify this constraint using the following inequalities:
$\overrightarrow{a'd'} \times \overrightarrow{a'b'} > 0$,
$\overrightarrow{b'a'} \times \overrightarrow{b'c'} > 0$,
$\overrightarrow{c'b'} \times \overrightarrow{c'd'} > 0$,
$\overrightarrow{d'c'} \times \overrightarrow{d'a'} > 0$,
where $\times$ represents the cross product.
If one of the inequalities is not satisfied, the estimated homography is discarded, suppressing false positives in our framework.

Even after the above convexity check, we empirically found some false positives with moderate scores.
These were caused by a characteristic of the ORB feature;
the multi-scale FAST detector used in ORB tends to detect features at almost the same position (often at corners) with different scales.
These features are frequently considered as inliers at the same time, increasing scores between unrelated image pairs.
In order to reduce this phenomenon, we discard the inliers if the inliers have positions of both reference and query features closer than five pixels, reducing the scores of non-informative matches.

\section{Experimental evaluation}
In the experiments, the Stanford mobile visual search dataset\footnote{\url{https://sites.google.com/site/chenmodavid/mobile-visual-search}} is used.
It contains eight classes of images:
camera-phone images of books, business cards, CDs, DVDs, outdoor landmarks, museum paintings, text documents, and video clips.
Each class consists of 100 reference images and 400 query images.
As an indicator of retrieval performance, mean average precision (MAP; higher is better)~\cite{jeg_ijcv10} is used.
We adopt the ORB feature~\cite{rub_iccv11} implemented in the OpenCV library\footnote{\url{http://opencv.org/}}, where at most 900 features are extracted from four scales on average.
The number of VWs is fixed at 1024 in all methods and experiments.
The VWs and substring dictionary are trained using the MIR Flickr collection\footnote{\url{http://press.liacs.nl/mirflickr/}}.

\begin{table*}[tb]
	\centering
	\caption{Comparison of the proposed method with conventional methods in terms of substring extraction.
	For each method, MAP scores of eight classes are shown.}
	\label{tab:result1}
	\begin{tabular}{c|>{\centering\arraybackslash}p{1.2cm}>{\centering\arraybackslash}p{1.2cm}>{\centering\arraybackslash}p{1.2cm}>{\centering\arraybackslash}p{1.2cm}>{\centering\arraybackslash}p{1.2cm}>{\centering\arraybackslash}p{1.2cm}>{\centering\arraybackslash}p{1.2cm}>{\centering\arraybackslash}p{1.2cm}|>{\centering\arraybackslash}p{2cm}} \hline
			&book	&cards	&cd &dvd	&landmarks	&paintings	&text	&video	&average \\ \hline
BoBW~\cite{gal_iros11}		&0.610	&0.173	&0.427	&0.465	&0.080	&0.486	&0.125	&0.584	&0.369${\pm}$0.012 \\
\cite{zho_mm12}			&0.874	&0.463	&0.752	&0.811	&0.197	&0.671	&0.423	&0.824	&0.627${\pm}$0.020 \\
PROP		&\textbf{0.916}	&\textbf{0.535}	&\textbf{0.807}	&\textbf{0.897}	&\textbf{0.253}	&\textbf{0.718}	&\textbf{0.542}	&\textbf{0.853}	&\textbf{0.690}${\pm}$0.015 \\ \hline
	\end{tabular} \\
\end{table*}

\begin{figure}[tb]
	\centering
	\includegraphics[width=0.8\linewidth]{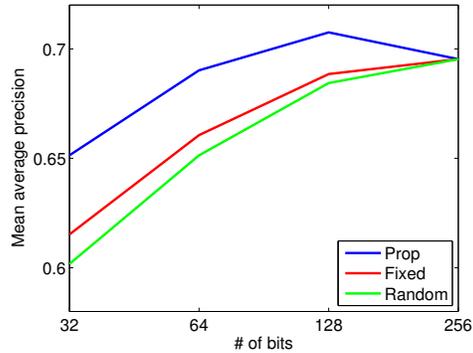}
	\caption{Average MAP scores of eight classes as a function of the length of substrings $T$.}
	\label{fig:bits}
\end{figure}

\subsection{Effect of substring generation}
First, the effectiveness of the proposed substring generation method was evaluated.
Table~\ref{tab:result1} summarizes the experimental results.
The MAP scores of eight classes are shown for three methods:
(1) BoBW~\cite{gal_iros11},
(2) Zhou's method~\cite{zho_mm12},
where we used the first 10 bits to define 1024 ($=2^{10}$) VWs and used the next 64 bits as the substring, and (3) the proposed method with 64 bits as the substring ($T = 64$).
In order to focus on the evaluation of the substring generation methods, conventional tf-idf scoring~\cite{low04} is used for all methods.
Comparing BoBW with Zhou's method, we can see that the use of the substring improves accuracy dramatically.
However, the proposed method can further improve accuracy by adaptively generating the substring.

For a statistical test, we split queries into five disjoint sets and calculated the average MAP for each set.
In Table~\ref{tab:result1}, the standard deviation of the average MAP is shown.
Furthermore, we conducted a paired-samples t-test where the null hypothesis is that there is no difference between the average MAPs of \cite{zho_mm12} and PROP. The two-tailed $P$-value for this hypothesis was 0.0001 and the null hypothesis was rejected.
Therefore, we can say that the improvement is statistically significant.

Next, the effect of the number of selected bits and the selection of methods are evaluated.
Here, in addition to the proposed method, we evaluate two selection methods: \textsf{Fixed} and \textsf{Random}. \textsf{Fixed} is a modified version of the proposed method,
where a substring is created using the first fixed $T$ bits in all visual words.
\textsf{Random} uses $T$ bits randomly selected for each visual word.
These three selection methods become identical when $T = 256$, where full binary strings of ORB features are always used.
Figure~\ref{fig:bits} shows the average MAP scores of eight classes as a function of the length of substrings $T$, comparing these three selection methods.
We can see that the proposed method achieves the best scores among these variants for all $T$.
\textsf{Fixed} is slightly better than \textsf{Random}.
This is reasonable because, in the ORB algorithm, binary tests are sorted according to their entropies;
the leading bits are more informative.
The interesting observation is that the proposed method at $T = 128$ outperforms the proposed method at $T = 256$, while a longer bit string achieves a better result using the \textsf{Fixed} and \textsf{Random} methods.
This implies that using non-informative or correlated bits degrades search accuracy.

\begin{table*}[tb]
	\centering
	\caption{Comparison of the proposed method with conventional methods in terms of scoring.}
	\label{tab:result2}
	\begin{tabular}{c|>{\centering\arraybackslash}p{1.2cm}>{\centering\arraybackslash}p{1.2cm}>{\centering\arraybackslash}p{1.2cm}>{\centering\arraybackslash}p{1.2cm}>{\centering\arraybackslash}p{1.2cm}>{\centering\arraybackslash}p{1.2cm}>{\centering\arraybackslash}p{1.2cm}>{\centering\arraybackslash}p{1.2cm}|>{\centering\arraybackslash}p{2cm}} \hline
			&book	&cards	&cd &dvd	&landmarks	&paintings	&text	&video	&average \\ \hline
PROP+GW		&0.943	&0.602	&0.849	&0.930	&0.278	&0.740	&0.568	&0.900	&0.726${\pm}$0.017 \\
PROP+LNo		&0.927	&0.515	&0.830	&0.924	&0.282	&0.758	&0.501	&0.909	&0.706${\pm}$0.008 \\
PROP+LNm	&\textbf{0.955}	&\textbf{0.609}	&\textbf{0.873}	&\textbf{0.944}	&\textbf{0.289}	&\textbf{0.773}	&\textbf{0.570}	&\textbf{0.914}	&\textbf{0.741}${\pm}$0.010 \\ \hline
	\end{tabular} \\
\end{table*}

\subsection{Comparison in scoring function}
Second, we evaluated the proposed scoring method.
The proposed adaptive substring extraction method ($T = 64$) with Gaussian weighting (\textsf{GW})
\footnote{
For the Gaussian weighting,
we set $\sigma = 9$,
which achieved the best performance in our preliminary experiments,
while $\sigma = 16$ in \cite{jeg_cvpr09}.
}
is used as a conventional method.
From Table~\ref{tab:result2}, it is shown that, while the original LN scoring method (\textsf{LNo}) is inferior to \textsf{GW}, the proposed modified LN scoring method (\textsf{LNm}) outperforms GW by 1.5\% and the method in \cite{zho_mm12} by 11\% in MAP.
This is because scores of original LN are directly affected by the density of feature vectors;
the score of a query feature in dense space tends to be low, while the score of a query feature in sparse space tends to be high.
The proposed scoring method normalizes the score using the distance between the query feature and its $K$-th nearest neighbor feature in the database.
Thus, all query features can equally contribute to the similarity score, improving the final result.
As the overhead of the proposed method is negligible, the proposed system can improve retrieval accuracy with the same memory requirements and almost the same computational cost as conventional methods.
We also conducted the paired-samples t-test as done in Section 5.1.
The two-tailed p-value was 0.003 when the average MAPs of \textsf{GW} and \textsf{LNm} are assumed to be the same.
Thus, it can be said that the proposed scoring method achieves a statistically significant improvement in term of MAP.

\begin{figure}[tb]
	\centering
	\includegraphics[width=0.8\linewidth]{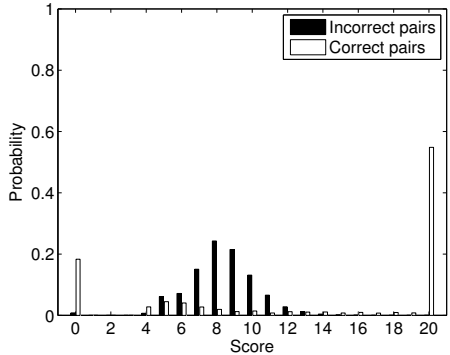} \\
	(a) GV \\
	\includegraphics[width=0.8\linewidth]{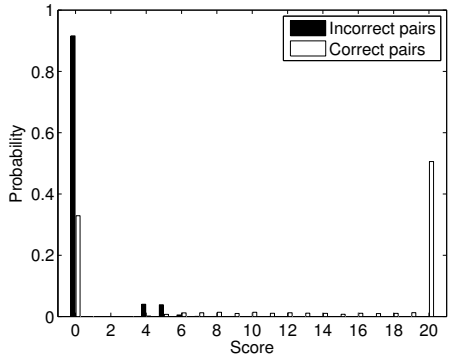} \\
	(b) GV+CC \\
	\caption{Distributions of the correct and incorrect image pairs with and without the convexity check.
	Scores larger than 20 are merged into the bin with 20 score.}
	\label{fig:gv}
\end{figure}

\begin{figure}[tb]
	\centering
	\includegraphics[width=\linewidth]{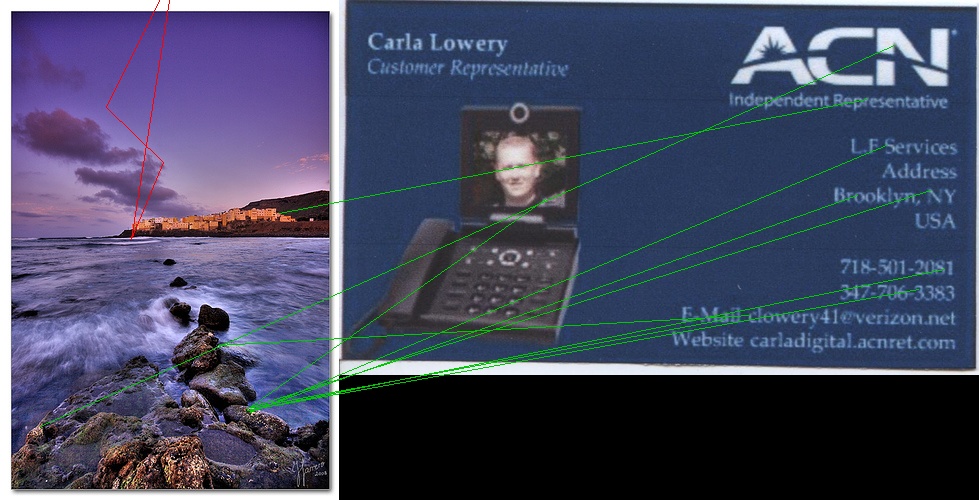} \\
	\includegraphics[width=\linewidth]{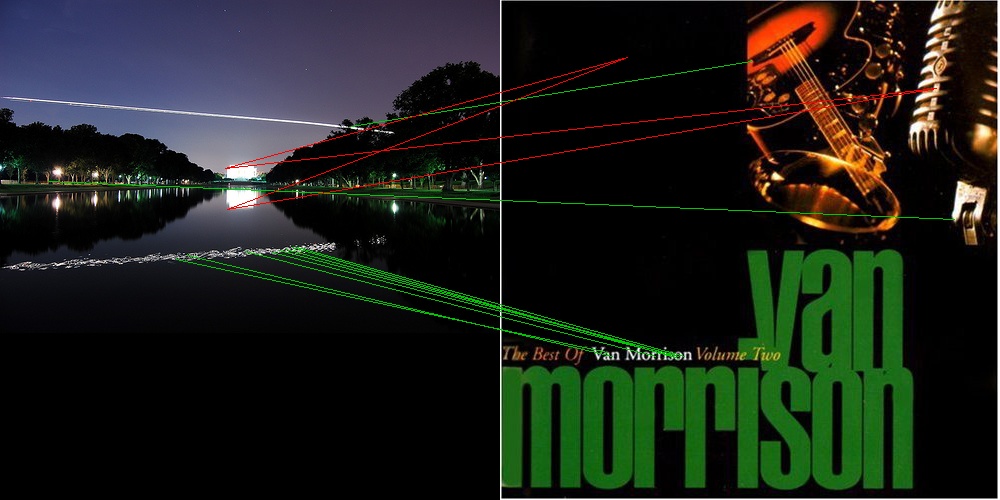} \\
	\includegraphics[width=\linewidth]{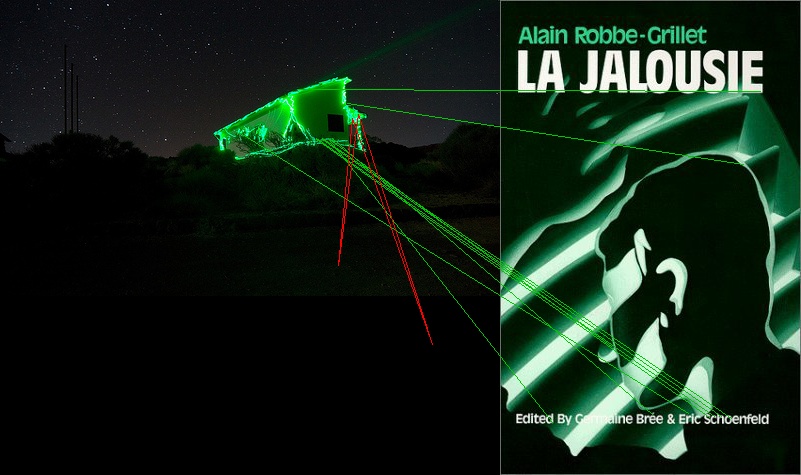} \\
	\caption{False positive examples after geometric verification.
	For each row, the left (resp. right) image represents the query (resp. reference) image.
	Green lines represent false inliers and red quadrilateral represents four sides of the reference image projected to the query image by the estimated Homography matrix.}
	\label{fig:gv_fp}
\end{figure}

\begin{figure}[tb]
	\centering
	\includegraphics[width=0.8\linewidth]{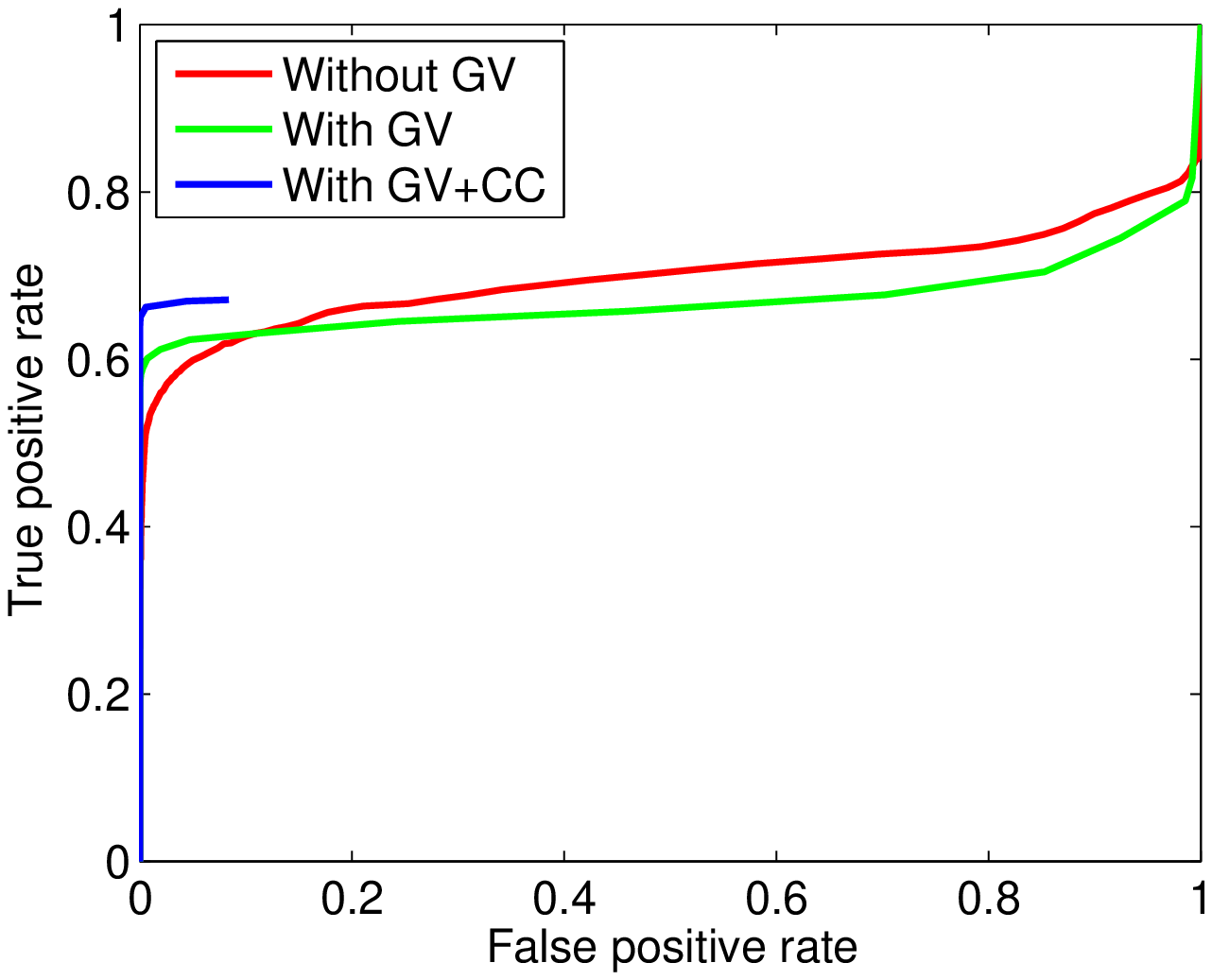}
	\caption{Receiver Operating Characteristic (ROC) curves for the methods without geometric verification, with geometric verification, and with geometric verification and convexity check.}
	\label{fig:roc}
\end{figure}

\subsection{Results with geometric verification}
\label{sec:gvresult}
The accuracy of the proposed framework with geometric verification is evaluated.
In this study, we perform geometric verification against the top three results of \textsf{LNm}.
First, we evaluate the effectiveness of the convexity check method described in Section~\ref{sec:gv} against false positive suppression.
For this purpose, we conducted an experiment where 10,000 images distinct from the reference images of the eight classes are used as queries and the databases of the eight classes are searched, obtaining 80,000 search results in total.
By taking the top-1 scores from these results, we can estimate the distribution of the scores between incorrect image pairs.
We also estimate the distribution of the scores between correct image pairs using the queries and references of eight classes.
The distributions of the correct and incorrect image pairs obtained by standard geometric verification (\textsf{GV}) and by the convexity check described in Section~\ref{sec:gv} (\textsf{GV+CC}) are shown in Figure~\ref{fig:gv} (a) and (b) respectively.
We can see that standard geometric verification returns relatively high scores even for incorrect image pairs.
In contrast, geometric verification with the convexity check can suppress most of the false positives.
Figure~\ref{fig:gv_fp} shows false positive examples after geometric verification, where red quadrilateral represents four sides of the reference image projected to the query image by the estimated Homography matrix.
We can see that the red quadrilaterals are complex (self-intersecting) and they can be filtered out by the proposed convexity check process.

Figure~\ref{fig:roc} shows the Receiver Operating Characteristic (ROC) curves for the methods without geometric verification, with geometric verification, and with geometric verification and the convexity check.
It can be said that both geometric verification and convexity check are vital to improving the accuracy under the constraint of a low false positive rate.
Table~\ref{tab:result3} shows the detection accuracy under the constraint that the false positive rate becomes zero.
Without geometric verification, it is impossible to achieve both reasonable detection accuracy and zero false positive rate.
Comparing \textsf{GV}/\textsf{GV+CC} with \textsf{LNm} in Table~\ref{tab:result2}, accuracy declines because geometric verification sometimes discards true positives in order to achieve a zero false positive rate.
The degradation of accuracy caused by GV is particularly prominent in the landmark and painting classes.
The landmark class includes many non-planar objects, and GV based on homography sometimes fails.
The painting class includes fewwer textured images, and lacks enough inliers in RANSAC.

\begin{table*}[tb]
	\centering
	\caption{Detection accuracy under the constraint that false positive rate becomes zero.}
	\label{tab:result3}
	\begin{tabular}{c|>{\centering\arraybackslash}p{1.2cm}>{\centering\arraybackslash}p{1.2cm}>{\centering\arraybackslash}p{1.2cm}>{\centering\arraybackslash}p{1.2cm}>{\centering\arraybackslash}p{1.2cm}>{\centering\arraybackslash}p{1.2cm}>{\centering\arraybackslash}p{1.2cm}>{\centering\arraybackslash}p{1.2cm}|>{\centering\arraybackslash}p{1.2cm}} \hline
			&book	&cards	&cd		&dvd	&landmarks	&paintings	&text	&video	&average \\ \hline
Without GV	& 0.062 & 0.010 & 0.088 & 0.120 & 0.000 & 0.214 & 0.000 & 0.300 & 0.099 \\
With GV		& 0.826 & 0.380 & 0.690 & 0.813 & 0.018 & 0.607 & 0.370 & 0.765 & 0.559 \\
With GV+CC	& 0.868 & 0.505 & 0.790 & 0.893 & 0.058 & 0.654 & 0.473 & 0.818 & 0.632 \\ \hline
	\end{tabular} \\
\end{table*}

\subsection{Computational cost and memory requirement}
Table~\ref{tab:timing} shows processing times for feature extraction (\textsf{Feature}), quantization (\textsf{Quantize}), Hamming distance calculation and voting (\textsf{Hamming}), and geometric verification (\textsf{GV}).
These durations are measured on a standard PC with a Core i7 2600 CPU and 32 GB of main memory (\textsf{PC}), and on an IS11S smartphone (released in 2011) with Qualcomm Snapdragon S2 MSM8655 1 GHz (\textsf{SP}).
Extracting substrings is quite simple and thus fast, so the computational cost is negligible.
The times required for \textsf{Feature} and \textsf{GV} take significantly longer on the smartphone, while those for \textsf{Quantize} and \textsf{Hamming} are not.
This is because, in our implementation, \textsf{Quantize} and \textsf{Hamming} processes can be sped up using the ARM NEON SIMD operation on the smartphone.
We can see that recognition rates are about 14 fps on the PC and about 2 fps on the smartphone, achieving a real-time local MVS with no false positives.
Comparing our method and Zhou's method~\cite{zho_mm12}, Zhou's method is a little faster because Zhou's method does not require the quantization process in Table~\ref{tab:timing}.
However, the above acceleration results in non-optimal performance in terms of search precision as discussed in Section~\ref{sec:extended} and shown in Table~\ref{tab:result1}.

In terms of memory requirements, we assume that 100 images are indexed,  900 features are extracted per image, the number of VWs is 1024, and the length of the substring is 64.
Under these settings, $100{\times}900{\times}14$ bytes = 1.26 MB is required to store features in the inverted index, $1024{\times}32$ bytes {$\simeq$} 0.03MB for VWs,
and $1024{\times}64$ bytes {$\simeq$} 0.07 MB for the substring dictionary.
This amount of data can reasonably be included in application binary files such as .apk for Android or .ipa for iPhone.
For instance, the size of our the test application apk that includes all of the above data was 3.66 MB.
Comparing our method and Zhou's method~\cite{zho_mm12}, Zhou's method does not need to store VWs (0.03 MB) and the substring dictionary (0.07 MB).
This memory footprint is relatively small compared with that of the inverted index, and it does not increase even if the number of images is increased.

\begin{table}[tb]
	\centering
	\caption{Processing times [sec] of the proposed MVS system.}
	\label{tab:timing}
	\begin{tabular}{c|>{\centering\arraybackslash}p{1cm}>{\centering\arraybackslash}p{1cm}>{\centering\arraybackslash}p{1cm}>{\centering\arraybackslash}p{1cm}|>{\centering\arraybackslash}p{1cm}} \hline
			&Feature	&Quantize	&Hamming	&GV		&Total \\ \hline
PC			&0.009		&0.015		&0.013		&0.035	&0.072 \\
SP		&0.184		&0.076		&0.053		&0.211	&0.524 \\ \hline
	\end{tabular} \\
\end{table}

\section{Conclusion}
In this paper, we proposed a stand-alone mobile visual search system based on binary features.
In our system, a VW-dependent substring extraction method and a new scoring method are used.
It is shown that the proposed system can improve retrieval accuracy with the same memory requirements as conventional methods.
Geometric verification using a constraint on the configuration of a transformed reference image achieved no false positive results.
In future work, we would like to confirm the scalability of our framework and apply it to a server-client system, where our substring method would be useful in reducing communication traffic between a mobile device and a search server.

{\small
\bibliographystyle{ieee}
\bibliography{refs}

\begin{thebibliography}{10}\itemsep=-1pt

\bibitem{ala_cvpr12}
A.~Alahi, R.~Ortiz, and P.~Vandergheynst.
\newblock Freak: Fast retina keypoint.
\newblock In {\em Proc. of CVPR}, pages 510--517, 2012.

\bibitem{alc_eccv12}
P.~Alcantarilla, A.~Bartoli, and A.~Davison.
\newblock Kaze features.
\newblock In {\em Proc. of ECCV}, 2012.

\bibitem{alc_bmvc13}
P.~Alcantarilla, J.~Nuevo, and A.~Bartoli.
\newblock Fast explicit diffusion for accelerated features in nonlinear scale
  spaces.
\newblock In {\em Proc. of BMVC}, 2013.

\bibitem{ara_cvpr13}
R.~Arandjelovi{\'c} and A.~Zisserman.
\newblock All about {VLAD}.
\newblock In {\em Proc. of CVPR}, 2013.

\bibitem{bay_cviu08}
H.~Bay, A.~Ess, T.~Tuytelaars, and L.~V. Gool.
\newblock Surf: Speeded up robust features.
\newblock {\em CVIU}, 110(3):346--359, 2008.

\bibitem{cal_eccv10}
M.~Calonder, V.~Lepetit, C.~Strecha, and P.~Fua.
\newblock Brief: Binary robust independent elementary features.
\newblock In {\em Proc. of ECCV}, pages 778--792, 2010.

\bibitem{dav_mm14}
D.~M. Chen and B.~Girod.
\newblock Memory-efficient image databases for mobile visual search.
\newblock {\em IEEE MultiMedia}, 21(1):14--23, 2014.

\bibitem{chu_cvpr05}
O.~Chum and J.~Matas.
\newblock Matching with prosac - progressive sample consensus.
\newblock In {\em Proc. of CVPR}, 2005.

\bibitem{chum_accv04}
O.~Chum, J.~Matas, and \v{S}. Obdr\v{z}\'{a}lek.
\newblock Enhancing ransac by generalized model optimization.
\newblock In {\em Proc. of ACCV}, 2004.

\bibitem{chu_iccv07}
O.~Chum, J.~Philbin, J.~Sivic, M.~Isard, and A.~Zisserman.
\newblock Total recall: {A}utomatic query expansion with a generative feature
  model for object retrieval.
\newblock In {\em Proc. of ICCV}, 2007.

\bibitem{gal_iros11}
D.~G{\'a}lvez-L{\'o}pez and J.~D. Tard{\'o}s.
\newblock Real-time loop detection with bags of binary words.
\newblock In {\em Proc. of IROS}, pages 51--58, 2011.

\bibitem{gio_vldb99}
A.~Gionis, P.~Indyk, and R.~Motwani.
\newblock Similarity search in high dimensions via hashing.
\newblock In {\em Proc. of VLDB}, pages 518--529, 1999.

\bibitem{hei_eccv12}
J.~Heinly, E.~Dunn, and J.-M. Frahm.
\newblock Comparative evaluation of binary features.
\newblock In {\em Proc. of ECCV}, pages 759--773, 2012.

\bibitem{jeg_cvpr09}
H.~J{\'e}gou, M.~Douze, and C.~Schmid.
\newblock On the burstiness of visual elements.
\newblock In {\em Proc. of CVPR}, pages 1169--1176, 2009.

\bibitem{jeg_ijcv10}
H.~J{\'e}gou, M.~Douze, and C.~Schmid.
\newblock Improving bag-of-features for large scale image search.
\newblock {\em IJCV}, 87(3):316--336, 2010.

\bibitem{jeg10}
H.~J{\'e}gou, M.~Douze, and C.~Schmid.
\newblock Product quantization for nearest neighbor search.
\newblock {\em TPAMI}, 33(1):117--128, 2011.

\bibitem{jeg_cvpr10}
H.~J{\'e}gou, M.~Douze, C.~Schmid, and P.~P{\'e}rez.
\newblock Aggregating local descriptors into a compact image representation.
\newblock In {\em Proc. of CVPR}, pages 3304--3311, 2010.

\bibitem{jeg_pami12}
H.~J{\'e}gou, F.~Perronnin, M.~Douze, J.~S{\'a}nchez, P.~P{\'e}rez, and
  C.~Schmid.
\newblock Aggregating local image descriptors into compact codes.
\newblock {\em TPAMI}, 34(9):1704--1716, 2012.

\bibitem{leu_iccv11}
S.~Leutenegger, M.~Chli, and R.~Siegwart.
\newblock Brisk: Binary robust invariant scalable keypoints.
\newblock In {\em Proc. of ICCV}, pages 2548--2555, 2011.

\bibitem{lev_wacv16}
G.~Levi and T.~Hassner.
\newblock Latch: Learned arrangements of three patch codes.
\newblock In {\em Proc. of WACV}, 2016.

\bibitem{low04}
D.~G. Lowe.
\newblock Distinctive image features from scale-invariant keypoints.
\newblock {\em IJCV}, 60(2):91--110, 2004.

\bibitem{mcc_cvpr12}
S.~McCann and D.~G. Lowe.
\newblock Local naive bayes nearest neighbor for image classification.
\newblock In {\em Proc. of CVPR}, 2012.

\bibitem{mik_pami05}
K.~Mikolajczyk and C.~Schmid.
\newblock A performance evaluation of local descriptors.
\newblock {\em TPAMI}, 27(10):1615--1630, Oct. 2005.

\bibitem{mik_ijcv05}
K.~Mikolajczyk, T.~Tuytelaars, C.~Schmid, A.~Zisserman, J.~Matas,
  F.~Schaffalitzky, T.~Kadir, and L.~V. Gool.
\newblock A comparison of affine region detectors.
\newblock {\em IJCV}, 60(1-2):43--72, Nov. 2005.

\bibitem{mik_eccv10}
A.~Mikul\'{\i}k, M.~Perdoch, O.~Chum, and J.~Matas.
\newblock Learning a fine vocabulary.
\newblock In {\em Proc. of ECCV}, pages 1--14, 2010.

\bibitem{mik_ijcv13}
A.~Mikul\'{\i}k, M.~Perdoch, O.~Chum, and J.~Matas.
\newblock Learning vocabularies over a fine quantization.
\newblock {\em IJCV}, 103(1):163--175, 2013.

\bibitem{muj_crv12}
M.~Muja and D.~G. Lowe.
\newblock Fast matching of binary features.
\newblock In {\em Proc. of CRV}, 2012.

\bibitem{nis06}
D.~Nist{\'e}r and H.~Stew{\'e}nius.
\newblock Scalable recognition with a vocabulary tree.
\newblock In {\em Proc. of CVPR}, pages 2161--2168, 2006.

\bibitem{pan_iccv13}
J.~Panda, M.~S. Brown, and C.~V. Jawahar.
\newblock Offline mobile instance retrieval with a small memory footprint.
\newblock In {\em Proc. of ICCV}, 2013.

\bibitem{per_eccv10}
F.~Perronnin, J.~S{\'a}nchez, and T.~Mensink.
\newblock Improving the fisher kernel for large-scale image classification.
\newblock In {\em Proc. of ECCV}, pages 143--156, 2010.

\bibitem{phi07}
J.~Philbin, O.~Chum, M.~Isard, J.~Sivic, and A.~Zisserman.
\newblock Object retrieval with large vocabularies and fast spatial matching.
\newblock In {\em Proc. of CVPR}, pages 1--8, 2007.

\bibitem{phi_cvpr08}
J.~Philbin, O.~Chum, M.~Isard, J.~Sivic, and A.~Zisserman.
\newblock Lost in quantization: Improving particular object retrieval in large
  scale image databases.
\newblock In {\em Proc. of CVPR}, pages 1--8, 2008.

\bibitem{ros_iccv05}
E.~Rosten and T.~Drummond.
\newblock Fusing points and lines for high performance tracking.
\newblock In {\em Proc. of ICCV}, pages 1508--1515, 2005.

\bibitem{rub_iccv11}
E.~Rublee, V.~Rabaud, K.~Konolige, and G.~Bradski.
\newblock Orb: An efficient alternative to sift or surf.
\newblock In {\em Proc. of ICCV}, pages 2564--2571, 2011.

\bibitem{siv03}
J.~Sivic and A.~Zisserman.
\newblock Video google: A text retrieval approach to object matching in videos.
\newblock In {\em Proc. of ICCV}, pages 1470--1477, 2003.

\bibitem{spy_tmm14}
E.~Spyromitros-Xioufis, S.~Papadopoulos, I.~Kompatsiaris, G.~Tsoumakas, and
  I.~Vlahavas.
\newblock A comprehensive study over vlad and product quantization in
  large-scale image retrieval.
\newblock {\em TMM}, 16(6):1713--1728, 2014.

\bibitem{uch_acpr13}
Y.~Uchida and S.~Sakazawa.
\newblock Image retrieval with fisher vectors of binary features.
\newblock In {\em Proc. of ACPR}, 2013.

\bibitem{uch_gcce14}
Y.~Uchida, S.~Sakazawa, and S.~Satoh.
\newblock Binary feature-based image retrieval with effective indexing and
  scoring.
\newblock In {\em Proc. of GCCE}, 2014.

\bibitem{yan_ismar12}
X.~Yang and K.~Cheng.
\newblock Ldb: An ultra-fast feature for scalable augmented reality on mobile
  devices.
\newblock In {\em Proc. of ISMAR}, pages 49--57, 2012.

\bibitem{yan_pami14}
X.~Yang and K.~Cheng.
\newblock Local difference binary for ultra-fast and distinctive feature
  description.
\newblock {\em TPAMI}, 36(1), 2014.

\bibitem{zel_nips04}
L.~Zelnik-Manor and P.~Perona.
\newblock Self-tuning spectral clustering.
\newblock In {\em Proc. of NIPS}, pages 1601--1608, 2004.

\bibitem{zho_mm12}
W.~Zhou, Y.~Lu, H.~Li, and Q.~Tian.
\newblock Scalar quantization for large scale image search.
\newblock In {\em Proc. of MM}, 2012.

\end{thebibliography}
}

\end{document}